\pdfoutput=1

\documentclass[11pt,a4paper]{article}

\usepackage{acl}

\usepackage{times}
\usepackage{latexsym}

\usepackage[T1]{fontenc}

\usepackage[utf8]{inputenc}

\usepackage{microtype}

\usepackage{hyperref}

\usepackage{algorithm}
\usepackage{algpseudocode}
\usepackage{graphicx}
\usepackage{enumitem}

\usepackage{comment}
\usepackage{soul}
\usepackage{xcolor}

\title{Improving Gender Translation Accuracy with Filtered Self-Training}

\author{Prafulla Kumar Choubey\thanks{~~Equal contribution.} \thanks{~~Work done as an intern at Amazon AI.}\\
  Texas A\&M University \\
  {\small{\texttt{prafulla.choubey@tamu.edu}}} \\\And
  Anna Currey\footnotemark[1], \hfill ~Prashant Mathur, \hfill Georgiana Dinu\\
  Amazon AI \\
  {\small{\texttt{\{ancurrey,pramathu,gddinu\}@amazon.com}}} \\}

\date{}

\begin{document}
\maketitle
\begin{abstract}
Targeted evaluations have found that machine translation systems often output incorrect gender, even when the gender is clear from context. Furthermore, these incorrectly gendered translations have the potential to reflect or amplify social biases. 

We propose a \textit{gender-filtered self-training} technique to improve gender translation accuracy on unambiguously gendered inputs. 
This approach uses a source monolingual corpus and an initial model to generate gender-specific pseudo-parallel corpora which are then added to the training data. 
We filter the gender-specific corpora on the source and target sides to ensure that sentence pairs contain and correctly translate the specified gender. 
We evaluate our approach on translation from English into five languages, finding that our models improve gender translation accuracy without any cost to generic translation quality. 
In addition, we show the viability of our approach on several settings, including re-training from scratch, fine-tuning, controlling the balance of the training data, forward translation, and back-translation. 

\end{abstract}


\section{Introduction} \label{sec:intro}
Recent work has drawn attention to the harms that machine learning algorithms can cause by reflecting or even amplifying data biases against protected groups~\citep{barocas-hardt-narayanan,10.5555/3379082}.
For the most part, machine translation (MT) studies on bias have focused on gender bias in neural machine translation (NMT) and have identified, using the taxonomy of \citet{blodgett-etal-2020-language}, a series of representational harms and stereotyping.\footnote{Representational harms occur when the model's performance is lower on input data associated with a protected group as opposed to the other groups. Stereotyping occurs when a model's prediction reflects negative stereotypes, for example about a specific ethnicity, or other stereotypical correlations, for example between professions and gender.}
For example, on input sentences that are underspecified in terms of gender, models often default to masculine or gender-stereotypical outputs~\citep{cho-etal-2019-measuring,DBLP:journals/corr/abs-1809-02208}, which can have the effect of excluding female and non-binary people (e.g.\ the sentence \textit{I am a doctor} spoken by a woman may be translated incorrectly as \textit{I am a (male) doctor}).
Even on unambiguously gendered inputs, NMT models can exhibit poorer performance, in terms of overall quality or gender translation accuracy, on content referring to feminine participants than masculine~\citep{bentivogli-etal-2020-gender,stanovsky-etal-2019-evaluating}.


\begin{figure}[t]
    \centering
    \includegraphics[scale=0.4]{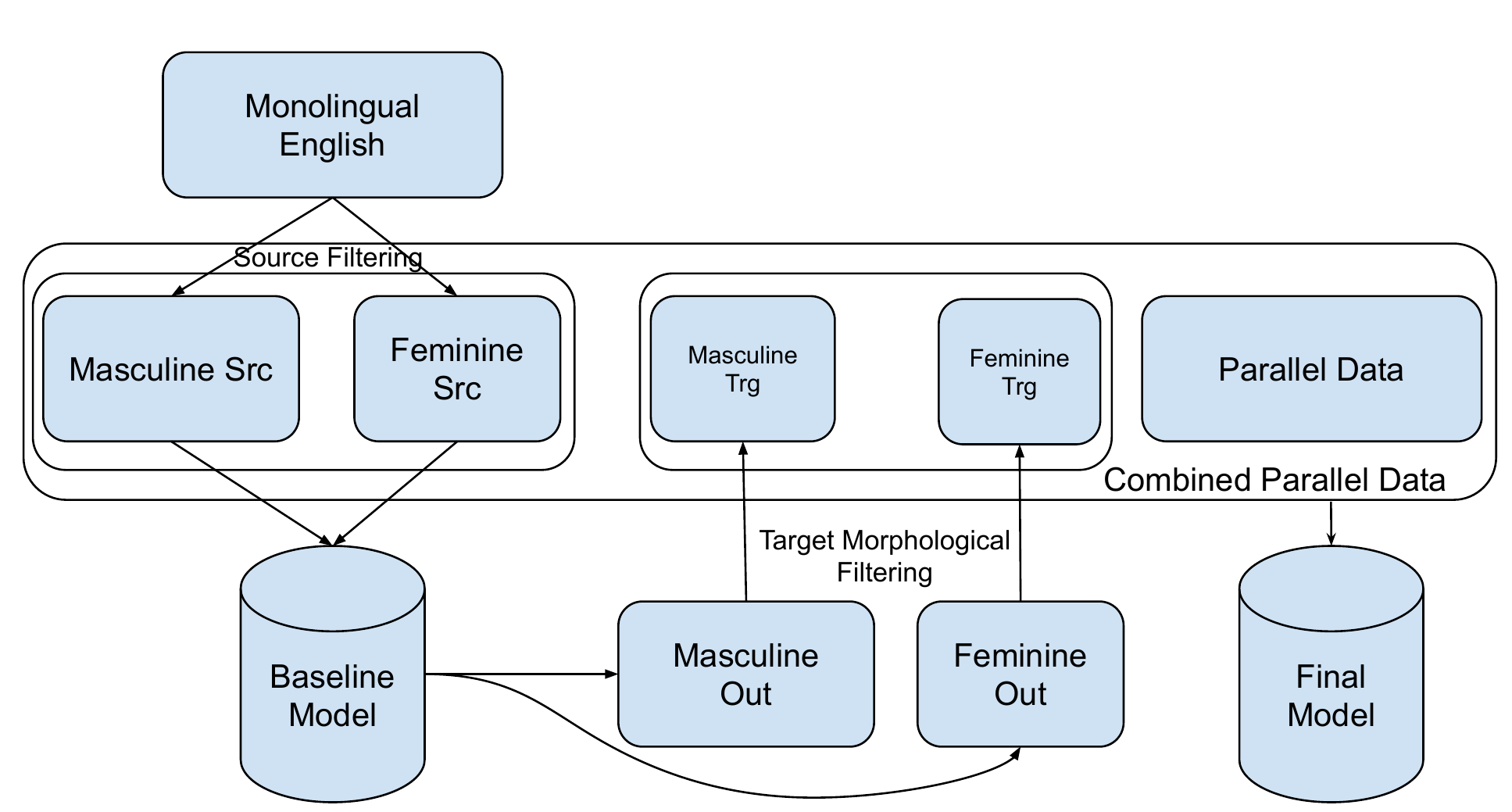}
    \caption{Gender-filtered self-training NMT approach.}
    \label{fig:method}
\end{figure}

In this paper, we take on the task of improving gender translation accuracy, focusing on unambiguous input where there is only one correct translation with respect to gender. This task is especially difficult when translating from languages without grammatical gender (e.g.\ English) into languages with extensive gender markings (e.g.\ German).


Known sources of gender bias in machine translation include selection bias (a.k.a.\ sample bias), which occurs when the input (source) distribution differs from that of the target application; label bias, which in MT occurs when gender-neutral sentences are translated predominantly into a specific gender or when the gender is translated incorrectly in the training data; and over-amplification, which is a property of the machine learning model~\citep{shah-etal-2020-predictive}.
We focus on addressing selection bias, starting from the observation that most MT training data is gender imbalanced (partly due to the fact that MT training data is largely sourced opportunistically).
Indeed, Table~\ref{intro-datadist} shows the relative proportion of masculine-referring vs.\ feminine-referring\footnote{This work helps to mitigate representational harms caused by low gender translation accuracy in machine translation systems. Since male and female genders have been the focus of most targeted machine translation gender bias evaluations, we focus on these two genders and as such do not mitigate representational harms against non-binary genders. See our impact statement in section~\ref{sec:impact} for a broader discussion.} sentences in our training data (extracted using the source filtering algorithm described in section~\ref{sec:st}).
Although the vast majority (over 90\%) of the data is not specific to entities of one gender (`Mix' line in Table~\ref{intro-datadist}), there are at least 2.6 times more masculine-specific than feminine-specific sentences in all of our training sets.
Similarly, \citet{vanmassenhove-etal-2018-getting} show that, across 10 languages, only 30\% of the Europarl data has female speaker gender.


In this paper, we devise a method to address selection bias using source monolingual data alone. We show that self-training using gender-balanced monolingual data together with a filtering technique to reduce error propagation leads to improved gender translation across several languages and data sets. Our framework is very simple and scales easily to any target language for which a morphological tagger is available.
\begin{table}[t]
\small
\center
\begin{tabular}{|l|ccccc|}
\hline
 & \textit{en-de}  & \textit{en-fr} & \textit{en-he} &  \textit{en-it} & \textit{en-ru}\\     \hline
Fem & 0.7\% & 0.9\% & 1.9\% & 1.9\% & 0.5\% \\
Msc & 4.8\% & 3.1\% & 4.9\% & 4.9\% & 2.4\% \\
Mix & 94.5\% & 96.0\% & 93.2\% &93.2\% & 97.1\% \\
\hline
\end{tabular}
\caption{Distribution of feminine (Fem), masculine (Msc), and mixed data in our parallel training data. Training data consists of WMT/IWSLT corpora and is described in detail in the appendix.}
\label{intro-datadist}
\end{table}

The main contributions of this paper are:
\begin{enumerate}[noitemsep]
    \item We propose a self-training technique that leverages naturally occurring monolingual corpora exhibiting diverse gender phenomena. This data is translated and filtered to create gender-balanced additional training data.
    \item We show that the method leads to significant improvements in accuracy of gender translation and improves performance on both feminine and masculine gendered input.
    \item We evaluate our approach on several languages pairs and in different experimental settings, showing that it can be adapted to a fine-tuning paradigm and to back-translation.
\end{enumerate}

\section{Gender-Filtered Self-Training}
\label{sec:st}

\begin{table*}[ht]
\begin{tabular}{l}
\textbf{Target-filtered sentences} \\ \hline
My \textbf{daughter} is hurt at being rejected by the \textbf{girl} \textbf{she} called \underline{\textbf{her} best friend}\\
Meine Tochter ist verletzt, weil sie von dem M{\"a}dchen, das sie als \underline{ihren besten Freund} [...]\\
\\
Another passenger was held for three days for using \underline{\textbf{her}} phone on board a flight [...]\\
\underline{Ein weiterer Passagier} wurde drei Tage lang festgehalten, weil \underline{er sein} Telefon [...]\\
\end{tabular}
\caption{Example sentences removed from the \textit{en-de} self-training corpus during target filtering. Both sentences are from the feminine set; they are removed because there are words in the target with masculine grammatical gender (which are \underline{underlined} along with their aligned source words). Note that the source sentences were selected in the source filtering phase due to the feminine words in \textbf{bold}.}
\label{filter-example-truenegative}
\end{table*}

In this paper, we propose an approach for improving gender translation accuracy on machine-translated outputs of unambiguously gendered input sentences. 
We use filtering and self-training in order to augment the data used to train the MT model. 
This approach is illustrated in Figure~\ref{fig:method}. 

Our method assumes access to a parallel corpus $\mathrm{D_{par}}$ and a monolingual source corpus $\mathrm{D_{src}}$. 
We first train an initial model $\mathrm{\Theta_{ini}}$ on $\mathrm{D_{par}}$. 
Due to the skewed gender representation of the training data (see Table~\ref{intro-datadist}), $\mathrm{\Theta_{ini}}$ may fail to use relevant gender cues from context, incorrectly translating gender-unmarked feminine words (such as \textit{friend} in the sentence \textit{She is my friend}) as masculine or vice versa. 
The extent of such errors can vary with the amount and quality of the training data, the domain of the data, or linguistic features of the languages. 
Nonetheless, we assume that our baseline models can render the correct gender for at least some inputs~\citep{escude-font-costa-jussa-2019-equalizing}. 

Therefore, we use $\mathrm{\Theta_{ini}}$ to generate translations for gender-specific sentences extracted from the monolingual data $\mathrm{D_{src}}$. This forward-translated data is subsequently filtered to ensure that the translations contain the gender of the source, balanced by gender, and used to augment the training data.

The full process is illustrated in Algorithm~\ref{ST-procedure}. Below, we describe each step in more detail.

\begin{algorithm}[H]
{{
\caption{Self-Training for MT with Gender-Filtered Data}\label{ST-procedure}
\begin{algorithmic}[1]
    \Require{Parallel and src mono data} $\mathrm{D_{par}}$, $\mathrm{D_{src}}$
    \State Train $\mathrm{\Theta_{ini}}$ on $\mathrm{D_{par}}$
    \State For $\mathrm{gen}$ in $\mathrm{\{fem,msc\}}$
    \State \hspace*{\algorithmicindent} $\mathrm{D_{src}^{gen}}$ $\gets$ \Call{FilterSrc}{$\mathrm{D_{src},gen}$} 
    \State \hspace*{\algorithmicindent} $\mathrm{D_{trg}^{gen}}$ $\gets$ Translate $\mathrm{D_{src}^{gen}}$ using $\mathrm{\Theta_{ini}}$
    \State \hspace*{\algorithmicindent} $\mathrm{D_{par}^{gen}}$ $\gets$ \Call{FilterTrg}{$\mathrm{D_{src}^{gen}}$, $\mathrm{D_{trg}^{gen}, gen}$}
    \State Train $\mathrm{\Theta_{fin}}$ on $\mathrm{D_{par}}$ $+$ $\mathrm{D_{par}^{fem}}$ $+$ $\mathrm{D_{par}^{msc}}$
    \State \Return $\mathrm{\Theta_{fin}}$
\end{algorithmic}
}}
\end{algorithm}

\paragraph{\textsc{FilterSrc}:}
From the source (in our case, English) monolingual corpus $\mathrm{D_{src}}$, we extract a masculine and a feminine subset of sentence candidates ($\mathrm{D_{src}^{msc}}$ and $\mathrm{D_{src}^{fem}}$, respectively) to use for forward translating. 
Specifically, given lists of feminine and masculine words, we consider a source sentence masculine if it meets all of the following criteria: 
\begin{enumerate}[noitemsep,nolistsep]
    \item Has at least one masculine pronoun
    \item Does not have any feminine pronouns
    \item Does not contain any feminine words
\end{enumerate}
We use an equivalent set of criteria to extract feminine sentence candidates from the data. To define gender-specific words, we use a list from \citet{zhao-etal-2018-gender}\footnote{Found at \url{https://github.com/uclanlp/corefBias/blob/master/WinoBias/wino/generalized_swaps.txt}.}, which contains a total of 104 pairs of words, such as \textit{brother/sister} or \textit{boy/girl}.

\paragraph{\textsc{FilterTrg}:}
Filtering on the target side of the data is done to exclude sentence pairs for which the model failed to preserve the gender of the source sentence. 
We perform morphological analysis on the translations $\mathrm{D_{trg}^{msc}}$ of $\mathrm{D_{src}^{msc}}$ and remove translations that include any feminine lexical forms, and similarly for the translations of $\mathrm{D_{src}^{fem}}$.\footnote{Target-side filtering  is entirely based on grammatical gender. 
Since the target languages in our experiments mark gender on inanimate objects, this step may exclude valid translations where the gender is correctly preserved. 
However, we prefer to keep a smaller set of high-confidence sentences in order to avoid introducing too much noise during self-training. 
We analyze this trade-off in section~\ref{sec:filter-analysis}. }
Table~\ref{filter-example-truenegative} shows examples of sentences that passed source filtering but were removed during target filtering. 

Note that the target filtering step suffices to generate the gender-specific sentence pairs. 
However, source filtering reduces computational cost by limiting the search space for the candidate gender-specific sentences and reduces the risk of introducing wrongly translated sentence pairs which may pass target filtering. 

\paragraph{Gender-Filtered Self-Training:}
After filtering masculine and feminine pseudo-parallel corpora, the larger of the two gender-specific datasets is sub-sampled in order to balance the pseudo-parallel data.\footnote{Although adding a balanced corpus to imbalanced data will reduce the imbalance, it will not result in a perfectly balanced training corpus. For future work we plan to increase the amount of data used in augmentation and investigate its impact on performance. Balancing the original training data through filtering or counterfactual training may be a viable alternative. However, our proposed method of generating balanced data scales more easily as we show that a precision-oriented filtering algorithm suffices.} 
Finally, the parallel corpus $\mathrm{D_{par}}$ is concatenated with the two pseudo-parallel corpora $\mathrm{D_{par}^{fem}}$ and $\mathrm{D_{par}^{msc}}$ and used to train a final MT model $\mathrm{\Theta_{fin}}$.

\section{Evaluation}

\subsection{Gender Accuracy on WinoMT}
We evaluate our models on the WinoMT~\citep{stanovsky-etal-2019-evaluating} gender-annotated test sets. 
WinoMT consists of 3888 English sentences taken from the Winogender \cite{rudinger-etal-2018-gender} and WinoBias \cite{zhao-etal-2018-gender} datasets. Each sentence contains a target occupation which lacks gender marking at the lexical level, such as \textit{salesperson}. The gender of the referent is implicitly defined by a coreferential pronoun in the sentential context, leading to sentences such as \textit{The \underline{salesperson} sold some books to the librarian because it was \underline{her} job}, where \textit{salesperson} is implicitly but unambiguously feminine. 
The dataset distinguishes between anti- and pro-stereotypical occupations, and contains 3648 sentences equally balanced between masculine and feminine as well as pro-stereotypical and anti-stereotypical occupations. Target occupations in the remaining 240 sentences are identified with neutral gender (e.g.\ \textit{The technician told \underline{someone} that \underline{they} could pay with cash}) and are excluded from the stereotype annotation. 

\paragraph{WinoMT Metrics:} On the WinoMT benchmark, the automated evaluation strategy first uses $fast\_align$ \cite{dyer-etal-2013-simple} to find the alignment for the target occupation in the translation. 
Then, using heuristic rules over language-specific morphological analysis, it identifies the gender of the translated occupation and uses three metrics to estimate the overall bias. 
\textbf{Accuracy} is the percentage of translations that reflect the correct gender of target occupation, while  $\mathbf{\Delta G}$ and $\mathbf{\Delta S}$ are defined as the difference in $F_1$ scores between masculine and feminine and between pro-stereotypical and anti-stereotypical target occupations respectively.

\paragraph{$\mathbf{\Delta R}$:}
$\Delta G$ 
may not give a complete picture of gender bias when the test set includes samples with unambiguously neutral gender (e.g. WinoMT sentences with \textit{they}). 
To understand  how this can happen, consider two hypothetical machine translation models, where both models have equal accuracy on feminine and masculine inputs but differ in how they treat neutral inputs.\footnote{The correct gender translations of such sentences depends on the grammatical conventions of the target language.} 
Model A translates all neutral inputs as masculine, whereas model B translates half of the neutral inputs as masculine and half as feminine. 
In this scenario, model A will have a lower $\Delta G$ (in particular, it will be negative) because it has lower precision on masculine inputs but the same recall for masculine and feminine inputs. 
However, we argue that model A may still be biased towards the masculine gender, since it defaults to masculine outputs when the inputs are neutral.

Therefore, we propose a new metric $\mathbf{\Delta R}$, defined as the difference in \textbf{recall} between masculine and feminine samples, to complement the existing metrics and give a more complete picture of model biases. 
$\Delta R$ decouples precision from the $\Delta G$ metric by excluding these neutral inputs from consideration. 
Thus, it is an indicator of the model's bias towards \textbf{outputting} masculine vs.\ feminine gender.

\paragraph{Human Evaluations:} The WinoMT benchmark was originally validated using human annotators, and the agreement between the annotators and the automatic gender accuracy metric was over 85\% across all language pairs and system outputs. 
We perform a similar human evaluation on select languages. Fluent speakers of German, Italian, and Russian were asked to annotate the gender translation accuracy of a random subset of 100 sentences from WinoMT, which was balanced for masculine/feminine and for pro-/anti-stereotype. 
Annotators were instructed to classify a translation as one of five labels. 
In addition to \textit{masculine} or \textit{feminine} (as in WinoMT automatic evaluations), we added the options \textit{inconsistent} (if some words in the translation indicated one gender and some indicated another), \textit{ambiguous}\footnote{Although we assume that the \textbf{input} sentences are unambiguous for gender, the outputs might still be ambiguous for gender. See Table~\ref{ambig-example} for an example.} (if the translation was perfectly valid for both masculine and feminine cases), and \textit{N/A} (if the entity of interest was omitted from the translation). 
This annotation was used to classify translations as \textit{incorrect} if they were inconsistent, N/A, or a different gender from the source (e.g.\ masculine if the source sentence was feminine), and \textit{correct} if they were ambiguous or the same gender as the source.

\subsection{Gender Accuracy on MuST-SHE}
In addition to the WinoMT benchmark, we also use the MuST-SHE gender-specific translation dataset~\citep{bentivogli-etal-2020-gender} to evaluate gender translation accuracy. 

MuST-SHE consists of roughly 1000 triplets of audio, transcript, and reference translations taken from MuST-C \cite{di-gangi-etal-2019-must} for \textit{en-fr} and \textit{en-it}. 
Each triplet is identified with either masculine or feminine gender based on speaker gender (category 1) or explicit gender markers such as pronouns (category 2). 
Furthermore, for each correct reference translation, the dataset includes a wrong alternative translation that changes the gender-marked words (feminine words are changed to masculine and vice versa). 
The dataset is balanced between masculine and feminine genders and between categories 1 and 2.

\paragraph{Automatic Metrics for MuST-SHE:} 
We use the category 2 samples (which contain explicitly marked gender words) from the MuST-SHE dataset to evaluate the overall translation quality for \textit{en-fr} and \textit{en-it}. 
Following \citet{bentivogli-etal-2020-gender}, we evaluate models using gender accuracy for translations on the correct references for both masculine and feminine samples.
We also look at $\Delta Acc$, which is the difference between the gender accuracy of translation with respect to correct and gender-swapped wrong references. 
For $\Delta Acc$, higher is better as this indicates that the model is closer to the correct reference than to the counterfactual reference.

\subsection{Generic Quality}
Our main goal is to improve gender translation accuracy. 
Additionally we measure generic quality using BLEU and human evaluations to investigate if the models lead to overall translation quality loss. Generic quality evaluations on the WinoMT test set also allow us test whether changes in gender accuracy are visible from a quality standpoint to speakers of the target languages.


\section{Experiments}
\label{sec:exp}

With source as English (EN), we experiment on five target languages from four families, all of which have grammatical gender: French (FR), Italian (IT), Russian (RU), Hebrew (HE), and German (DE). 
Our experiments include both low-resource and high-resource settings. 
Table~\ref{experiments-datasizes} shows the number of parallel training sentences after preprocessing, and the number of sentences in the pseudo-parallel corpus after source filtering and target filtering. 
For a full description of the data used, see Appendix~\ref{sec:app-params}. 

\begin{table}
\small
\center
\begin{tabular}{|l|ccccc|}
\hline
Dataset & \textit{en-de}  & \textit{en-fr} & \textit{en-he} &  \textit{en-it} & \textit{en-ru}\\     \hline
$\mathrm{D_{par}}$ & 5.2M & 35.7M & 180k & 161k & 1.6M\\
$\mathrm{D_{src}^{fem}}$ & 1.8M & 4.2M & 1.8M & 1.8M & 1.8M \\
$\mathrm{D_{par}^{fem}}$ & 428k & 150k & 29k & 81k & 184k\\
\hline
\end{tabular}
\caption{Number of sentences in each training set. $\mathrm{D_{par}}$ is the original parallel training data, $\mathrm{D_{src}^{fem}}$ is the source-filtered feminine monolingual data, and $\mathrm{D_{par}^{fem}}$ is the feminine data after target filtering. We downsample the larger masculine data to match the size of $\mathrm{D_{par}^{fem}}$.}
\label{experiments-datasizes}
\end{table}

We use Transformers \cite{vaswani2017attention} implemented in \textit{Fairseq-py} \cite{ott-etal-2019-fairseq}. 
Exact hyperparameters are detailed in the appendix. 

We experiment with the following models:
\begin{itemize}[noitemsep]
    \item \textbf{Baseline} models are trained on the original bitext $\mathrm{D_{par}}$ only; these correspond to $\mathrm{\theta_{ini}}$.
    \item \textbf{ST$\mathbf{_{Rand}}$} models are trained on $\mathrm{D_{par}}$ with additional data consisting of random pseudo-parallel sentence pairs.\footnote{Random pseudo-parallel sentence pairs for ST$_{Rand}$ are obtained through forward translation of the monolingual English corpora but without source and target side gender-based filtering. For fair comparison, we keep the size of random pairs equal to the combined size of masculine and feminine pairs used in ST$_{Gender}$.}
    \item \textbf{ST$\mathbf{_{Gender}}$} models are our proposed gender-filtered self-training models; they are trained on masculine and feminine pseudo-parallel data ($\mathrm{D_{par}^{msc}}$ and $\mathrm{D_{par}^{fem}}$) and on $\mathrm{D_{par}}$. 
    \item \textbf{+HD} models additionally use encoder sub-word embeddings that are hard-debiased following \citet{bolukbasi2016}.
\end{itemize}

\section{Results}
\label{sec:results}

\begin{table}
\small
\center
\begin{tabular}{|l|ccccc|}
\hline
Model & \textit{en-de}  & \textit{en-fr} & \textit{en-he} &  \textit{en-it} & \textit{en-ru}\\     \hline
Baseline & 41.7 & 40.5 & 23.4 & 34.5 & 25.7\\
 \ \ +HD & 41.8 & 40.7 & 23.5 & 34.4 & 25.5\\
ST$_{Rand}$ & \textbf{42.4} & \textbf{40.8} & \textbf{23.9} & 34.4 & \textbf{26.9} \\ \hline
ST$_{Gender}$ & 41.8 & 40.2 & 23.8 & \textbf{34.6} & 26.6 \\
 \ \ +HD & 42.0 & 40.4 & 23.8 & \textbf{34.6} & 26.5 \\
\hline
\end{tabular}
\caption{BLEU scores on generic test sets for the baselines and the proposed filtered self-training (ST$_{Gender}$) model. The test sets are from WMT for \textit{en-de}, \textit{en-fr}, and \textit{en-ru} and IWSLT for \textit{en-it} and \textit{en-he}, corresponding with the training data for each language pair.}
\label{result-bleu-main}
\end{table}

\subsection{Generic Quality}
\paragraph{Automatic Translation Quality:} 
We report case-sensitive de-tokenized BLEU using SacreBLEU \cite{post-2018-call}\footnote{SacreBLEU signature: BLEU+case.mixed+numrefs.1 +smooth.exp+tok.13a+version.1.4.10.} for all language pairs on the generic test sets (WMT or IWSLT test sets) in Table~\ref{result-bleu-main}. 
The results confirm that our proposed gender-filtered self-training procedure (ST$_{Gender}$) does not come at a trade-off in generic translation quality, compared to a baseline that does not use the gender-filtered pseudo-parallel data. 
We also observe an overall trend of small improvements from self-training models, irrespective of the data selection method. 

\paragraph{Human Quality Evaluations:} 
In order to better understand how the gender-filtered self-training data affects overall translation quality, we perform human evaluations of \textbf{quality} on a balanced, 300-sentence subset of the WinoMT test set. 
For each language pair, baseline
vs. ST$_{Gender}$ quality was evaluated on a six-point Likert scale by two professional translators. 
The quality scores, averaged between the two annotators, are shown in Table~\ref{result-human-quality}. 
For \textit{en-de}, \textit{en-he}, and \textit{en-it}, our ST$_{Gender}$ model improves significantly in overall quality. 
For \textit{en-fr} and \textit{en-ru}, there is no significant difference between the baseline and our model in terms of overall quality. 

\begin{table}[ht]
\small
\center
\begin{tabular}{|l|ccccc|}
\hline
Model & \textit{en-de}  & \textit{en-fr} & \textit{en-he} &  \textit{en-it} & \textit{en-ru}\\     \hline
Baseline & 4.52 & 4.55 & 2.84 & 3.50 & 3.86\\
ST$_{Gender}$ & \textbf{4.70} & 4.47 & \textbf{3.05} & \textbf{3.59} & 3.96\\
\hline
\end{tabular}
\caption{Human quality scores (average of two annotators) on a balanced subset of WinoMT. Differences outside the 95\% confidence interval are shown in \textbf{bold}.}
\label{result-human-quality}
\end{table}

\begin{table*}
\scriptsize
\center
\begin{tabular}{|l|ccc|ccc|ccc|ccc|ccc|c|}
\hline
&  \multicolumn{3}{|c|}{\textit{en-de}} & \multicolumn{3}{c|}{\textit{en-fr}} & \multicolumn{3}{c|}{\textit{en-he}} & \multicolumn{3}{c|}{\textit{en-it}}  & \multicolumn{3}{c|}{\textit{en-ru}} & \\ 
Model & Acc & $\Delta G$ & $\Delta R$ & Acc & $\Delta G$ &  $\Delta R$ &  Acc & $\Delta G$ &  $\Delta R$ &  Acc & $\Delta G$ &  $\Delta R$ &  Acc & $\Delta G$ & $\Delta R$ & Avg Acc\\     \hline
Baseline &  75.5 & {0.4} & 18.8  & 66.2 & {-0.2} & 13.3 & 47.5 & 13.8 & 30.9 & 38.8 & 31.5 & 50.9 & 34.1 & 32.7 & 46.6 & 52.4\\
 \ \ +HD & 75.4 & {0.4} & 19.2 & 66.1 & {0.2} & 15.3 & 47.7 & {12.5} & {28.3} & 39.1 & 32.4 & 52.3 & 34.3 & 31.7 & 45.7 & 52.5
 \\
ST$_{Rand}$ & 78.7 & -1.6 & 13.0 & 65.0 & 0.6 & 15.0 & 46.9 & 14.8 & 31.9 & 39.4 & 34.7 &  55.9 & 33.3 & 32.2 & {45.0}  & 52.7\\ \hline
ST$_{Gender}$ & \textbf{85.4} & -4.4 & -0.3 &  \textbf{71.0} & -1.3 & {10.2} &  48.6 & 13.8 & 31.6 & 50.0 & 18.0 & 41.3  &  39.0 & {30.3} & 47.8 & 58.8\\
 \ \ +HD & 85.3 & -4.4 & {-0.2} & 68.8 & 0.8 & 16.8 & \textbf{49.3} & 13.5 & 31.3 & \textbf{52.4} & {14.7} & {37.2}  & \textbf{39.4} & {30.3} & 47.8 & \textbf{59.0}\\
\hline
\end{tabular}
\caption{Performance of all systems on the WinoMT corpus using Accuracy (\textit{Acc}), $\Delta G$, and our proposed $\Delta R$. Comparison to other published results is difficult due to the differences in experimental settings. As a reference, \citet{stanovsky-etal-2019-evaluating} reports maximum accuracies of 74\% (\textit{en-de}), 63\% (\textit{en-fr}), 53\% (\textit{en-he}), 42\% (\textit{en-it}) and 39\% (\textit{en-ru}) using various commercial MT systems. \citet{saunders-byrne-2020-reducing} reports results of up to 81\% (\textit{en-de}) and 65\% (\textit{en-he}) for models not degrading generic quality, after fine-tuning on a handcrafted professions set and using lattice rescoring.}
\label{result-winomt-main}
\end{table*}

\subsection{Gender Translation Accuracy}
\paragraph{Automatic WinoMT Accuracy:} Table~\ref{result-winomt-main} compares all models on the WinoMT benchmark using accuracy (\textit{Acc}), $\Delta G$, and our newly proposed $\Delta R$ metric.\footnote{$\Delta S$ results are shown in the appendix, since debiasing according to stereotypes is not the main focus of this work.} 
The results show that our proposed method of self-training on gender-filtered data, ST$_{Gender}$, consistently yields large gains in accuracy (up to 11.2 points) over the baseline. 
Gains are largest for feminine inputs, although we see gains on masculine inputs as well. 
Full results for gender-specific $F_1$ are given in the appendix. 
By contrast, simply self-training on randomly sampled data (ST$_{Rand}$) does not improve gender accuracy significantly: average accuracy is 52.4 for the baseline and 52.7 for ST$_{Rand}$, indicating that gender filtering and balancing are important components of our approach. 

The ST$_{Gender}$ model also outperforms a baseline model that uses hard-debiasing~\cite{bolukbasi2016} on both accuracy and $\Delta R$ for all language pairs. 
Since hard-debiasing is orthogonal to the self-training methods, we also apply it to the  ST$_{Gender}$ model and observe small improvements on average. 
However, hard-debiased embeddings do not improve accuracy significantly on average for either the baseline model or our ST$_{Gender}$ model. 
Our findings are slightly different from those of \citet{escude-font-costa-jussa-2019-equalizing}, who found some small evidence for improved gender translation accuracy when using pre-trained hard-debiased embeddings on a different test set. 
On the other hand, \citet{gonen-goldberg-2019-lipstick} have also shown that hard-debiasing metrics may not meaningfully reduce gender bias. 
As such, and based on our results in Table~\ref{result-winomt-main}, we focus subsequent experiments on the simpler ST$_{Gender}$ models.

\begin{table}
\center
\begin{tabular}{|l|ccc|}
\hline
Model & \textit{en-de}  & \textit{en-it} & \textit{en-ru}\\     \hline
Baseline & 79\% & 50\% & 80\%\\
ST$_{Gender}$ & \textbf{93\%} & \textbf{65\%} & 79\%\\ \hline
\end{tabular}
\caption{Accuracy scores as evaluated by humans on a balanced subset of the WinoMT dataset. Scores for \textit{en-de} are averaged over two annotators.}
\label{result-winomt-human}
\end{table}

\paragraph{Human Accuracy Evaluations:} Table~\ref{result-winomt-human} shows the results for the human evaluations of gender accuracy on WinoMT. 
For \textit{en-de}\footnote{For \textit{en-de} we had two annotators, so we average their scores. Inter-annotator agreement was 78\% for the baseline and 97\% for ST$_{Gender}$.} and \textit{en-it}, we see a large increase in gender translation accuracy for our proposed ST$_{Gender}$ model compared to the baseline, while for \textit{en-ru}, there is no significant difference between the baseline and our proposed model. 
These scores largely corroborate the automatic WinoMT accuracy results in Table~\ref{result-winomt-main}, with larger differences in automatic scores corresponding to larger differences in human evaluation scores. 

\begin{table}[t]
\begin{tabular}{p{0.5cm}p{6.3cm}}
Src & \textbf{The guard} saved the editor from the criminals because \textbf{he} was on duty. \\ 
Trg &  \textbf{La guardia} ha salvato l'editore dai criminali perch{\'e} era in dovere.\\ 
\end{tabular}
\caption{Example of a WinoMT sentence and ambiguous output that was marked incorrect by WinoMT.}
\label{ambig-example}
\end{table}

Unlike standard WinoMT evaluations, we additionally allowed annotators to mark output genders as \textit{inconsistent} (which we mapped to incorrect) and \textit{ambiguous} (mapped to correct). 
Up to 19\% of the sentences in a given test set were marked as inconsistent, with baseline systems having slightly more inconsistent translations on average than ST$_{Gender}$ systems (12.8\% vs.\ 8.5\%). 
However, for all of the inconsistent sentences, automatic evaluations also marked those sentences as incorrect, since they all consisted of an incorrect (often stereotypical) gender on the noun but correct gender on a corresponding pronoun. 
Up to 11\% of the sentences in a given test set were marked as ambiguous -- cases where the gender of the given entity is not specified in the translation. 
Here, we did see some divergence from the WinoMT metric; Table~\ref{ambig-example} shows one such case. 
In the source sentence, the pronoun \textit{he} in the context indicates that the guard is male. 
In the translation, the only gendered word that refers to the guard is \textit{la guardia}, which, while grammatically feminine, could be applied to men or to women. 
Thus, the translation is ambiguous regarding the gender of the guard, although it is marked as incorrect by the automatic WinoMT evaluations. 

%
%


\paragraph{Automatic MuST-SHE Accuracy:}
In Table~\ref{result-mustshe-main}, we report accuracy for \textit{en-it} and \textit{en-fr} models on correct translations as well as the $\Delta Acc$ between correct and gender-swapped translations using category 2 data from the MuST-SHE corpus. 
For \textit{en-it}, our ST$_{Gender}$ model increases both accuracy and $\Delta Acc$ for feminine and masculine data. 
For the higher-resourced pair \textit{en-fr}, there is a small increase in accuracy and $\Delta Acc$ for feminine data, but also a (smaller) decrease in both metrics for masculine data. 

\begin{table*}[ht]
\center
\begin{tabular}{|l|cc|cc|cc|cc|}
\hline
&  \multicolumn{4}{|c|}{\textit{en-it}} & \multicolumn{4}{c|}{\textit{en-fr}} \\ 
&  \multicolumn{2}{|c|}{Fem} & \multicolumn{2}{c|}{Msc} & \multicolumn{2}{c|}{Fem} & \multicolumn{2}{c|}{Msc} \\ 
Model &  Acc  & $\Delta {Acc}$ & Acc  & $\Delta {Acc}$ & Acc  & $\Delta {Acc}$ & Acc  & $\Delta {Acc}$ \\     \hline 
Baseline & 32.5 & 2.5 & 58.8 & 48.8 & 57.5 & 46.0 &  \textbf{68.0} & \textbf{60.7}\\
ST$_{Gender}$ & \textbf{41.9} & \textbf{21.9} & \textbf{61.6} & \textbf{54.3} & \textbf{60.9} & \textbf{52.8} & 66.8 & 59.7\\
\hline
\end{tabular}
\caption{MuST-SHE performance measured in Accuracy (Acc) and $\Delta Accuracy$ ($\Delta Acc$).}
\label{result-mustshe-main}
\end{table*}

\section{Analysis}
\subsection{Self-Training from Scratch vs.\ Fine-Tuning}
\label{sec:abl-ft}

\begin{table}
\small
\center
\begin{tabular}{|l|ccccc|}
\hline
Model & \textit{en-de}  & \textit{en-fr} & \textit{en-he} &  \textit{en-it} & \textit{en-ru}\\     \hline
Baseline & 41.7 & 40.5 & 23.4 & 34.5 & 25.7\\
\hline
ST(R)$_{Gender}$ & \textbf{41.8} & 40.2 & \textbf{23.8} & \textbf{34.6} & \textbf{26.6} \\
ST(F)$_{Gender}$ & \textbf{41.8} & \textbf{41.1} & 23.6 & 34.3 & 24.6 \\ 
\hline
\end{tabular}
\caption{BLEU scores on the generic test data for the baseline model and the gender data augmented models: retrained (ST(R)$_{Gender}$) and fine-tuned (ST(F)$_{Gender}$).  
}
\label{result-bleu-ft}
\end{table}

\begin{table*}
\small
\center
\begin{tabular}{|l|cc|cc|cc|cc|cc|c|}
\hline 
&  \multicolumn{2}{|c|}{\textit{en-de}} & \multicolumn{2}{c|}{\textit{en-fr}} & \multicolumn{2}{c|}{\textit{en-he}} & \multicolumn{2}{c|}{\textit{en-it}}  & \multicolumn{2}{c|}{\textit{en-ru}} & \\ 
Model & Acc & $\Delta R$ & Acc & $\Delta R$ &  Acc & $\Delta R$ &  Acc &  $\Delta R$ &  Acc & $\Delta R$ & Avg Acc\\     \hline
Baseline &  75.5 & 18.8  & 66.2 & 13.3 & 47.5 & {30.9} & 38.8 & 50.9 & 34.1 & {46.6} & 52.4\\
 \hline
ST(R)$_{Gender}$ & \textbf{85.4} & {-0.3} &  {71.0} & {10.2} &  \textbf{48.6} & 31.6 & \textbf{50.0} & {41.3}  &  \textbf{39.0} & {47.8} & \textbf{58.8}\\
ST(F)$_{Gender}$ & 83.2 & 3.5 & \textbf{72.2} & {4.8} & 47.2 & 31.3 & 40.9 & 51.8 & 36.1 & 47.6 & 55.9\\ 
 \hline
\end{tabular}
\caption{Accuracy and $\Delta R$ scores on the WinoMT test data for the baseline model and the gender data augmented models: retrained (ST(R)$_{Gender}$) and fine-tuned (ST(F)$_{Gender}$). 
}
\label{result-winomt-ft}
\end{table*}

The main experiments (section~\ref{sec:results}) used models that were retrained (from scratch) using the additional filtered gendered data. 
We further explore the utility of the gender-filtered self-training data by fine-tuning existing models instead of retraining. Fine-tuning is advantageous because it is less costly.
In order to train these models, we fine-tune the baseline using the feminine and masculine samples, and additionally mix in an equal number of sentences from the original training corpus to avoid catastrophic forgetting (following the \textit{mixed fine-tuning} approach of \citealp{chu-etal-2017-empirical}). 

BLEU scores for baselines, retrained models, and fine-tuned models are shown in Table~\ref{result-bleu-ft}. 
For all language pairs except \textit{en-ru}, there is no significant drop in quality between the baseline and the fine-tuned models; for \textit{en-ru} there is a loss of about 1 BLEU.

Table~\ref{result-winomt-ft} shows WinoMT accuracy and $\Delta R$ results for the baseline, retrained models and fine-tuned models. 
On average,  the retrained models outperform the fine-tuned models. 
However, the fine-tuned models are consistently more accurate in gender translation than the baseline models, showing that fine-tuning is a viable low-cost alternative to retraining.

\subsection{Single-Gender Data Augmentation}
\label{sec:abl}
Although our original motivation (see Table~\ref{intro-datadist}) was to address gender imbalance in the training data, our proposed ST$_{Gender}$ models use gender-balanced augmented data, i.e.\ the same amount of feminine-specific and masculine-specific sentences in the self-training data. 

In this section we investigate the relative contribution of each augmentation corpus, by evaluating single-gender self-trained models: 

\begin{itemize}[noitemsep]
    \item \textbf{ST$_{Fem}$} models are trained on the original bitext $\mathrm{D_{par}}$ and on gender-filtered feminine sentence pairs $\mathrm{D_{par}^{fem}}$.
    \item \textbf{ST$_{Msc}$} models are trained on $\mathrm{D_{par}}$ and on downsampled gender-filtered masculine sentence pairs $\mathrm{D_{par}^{msc}}$.
\end{itemize}

In overall translation quality, all models perform similarly (see Appendix~\ref{sec:generic-test}). 
In Table~\ref{result-winomt-single}, we compare feminine-only, masculine-only, and joint self-training models to the baseline on the WinoMT benchmark using accuracy and $\Delta R$. 
As expected, ST$_{Fem}$ reduces the gap between recall for feminine and masculine samples, lowering $\Delta R$ by 3.6--19.8 points with respect to the baseline. At the same time, ST$_{Msc}$ increases $\Delta R$ overall, suggesting that gender-filtered self-training works as hypothesized and can be used to balance the training data distribution between masculine and feminine genders. 

\begin{table*}
\small
\center
\begin{tabular}{|l|cc|cc|cc|cc|cc|}
\hline 
&  \multicolumn{2}{|c|}{\textit{en-de}} & \multicolumn{2}{c|}{\textit{en-fr}} & \multicolumn{2}{c|}{\textit{en-he}} & \multicolumn{2}{c|}{\textit{en-it}}  & \multicolumn{2}{c|}{\textit{en-ru}} \\ 
Model & Acc & $\Delta R$ & Acc &  $\Delta R$ &  Acc &  $\Delta R$ &  Acc &  $\Delta R$ &  Acc & $\Delta R$\\     \hline
Baseline &  75.5 & 18.8  & 66.2 & 13.3 & 47.5 & 30.9 & 38.8 & 50.9 & 34.1 & 46.6\\
ST$_{Gender}$ & \textbf{85.4} & {-0.3} & \textbf{71.0} & 10.2 & 48.6 & 31.6 & \textbf{50.0} & 41.3 & \textbf{39.0} & 47.8\\ \hline
ST$_{Fem}$ &  84.0 & -1.0 & 69.3 & {5.6} & 48.2 & {25.1} &  48.0 & {24.6} & 37.1 & {43.0} \\
ST$_{Msc}$ & 75.5 & 22.0 & 64.8 & 25.0 & \textbf{48.8} & 34.7 & 40.1 & 65.0 & 34.8 & 50.9  \\
\hline
\end{tabular}
\caption{Accuracy and $\Delta R$ scores on the WinoMT test set. We compare the baseline model and our ST$_{Gender}$ model to models that use only feminine-specific additional data (ST$_{Fem}$) and masculine-specific additional data (ST$_{Msc}$).}
\label{result-winomt-single}
\end{table*}

On overall gender accuracy, ST$_{Fem}$ outperforms the baseline for all five language pairs and yields similar accuracy to the ST$_{Gender}$ model. 
On the other hand, ST$_{Msc}$ performs very closely to the baseline across all language pairs; this result for ST$_{Msc}$ could be due to the higher proportion of masculine than feminine samples in the original training data (see Table~\ref{intro-datadist}). 
The large improvements over the baseline for ST$_{Fem}$ compared to ST$_{Msc}$ highlight the under-representation of feminine-gender samples in the existing training corpora. 

ST$_{Gender}$, which is trained on both masculine and feminine additional data, outperforms ST$_{Fem}$ in accuracy but underperforms ST$_{Fem}$ in $\Delta R$. 
This is not surprising given that the ST$_{Fem}$ training data is more gender-balanced than the data used to train ST$_{Gender}$, which contains masculine pseudo-parallel data. 

\subsection{Self-Training vs.\ Back-Translation}
We have focused the experiemnts on a self-training approach using forward translation to generate gender-balanced data. 
However, if a large monolingual corpus is available in the target language, it is possible to extend the approach to back-translation (BT; \citealp{sennrich-etal-2016-improving}). 
This is done by applying the target filtering step\footnote{We used the additional target filtering criterion that the sentence have at least one pronoun.} on target-side monolingual data, using a baseline target$\rightarrow$source system to translate the filtered target data to the source language, and applying source filtering on the resulting pseudo-parallel source data. 

In order to verify this, we compare BT, ST, and an approach using both BT and ST for the \textit{en-de} setup. 
We use the same amount of pseudo-parallel data for both BT and ST (although the data itself is not the same, as it comes from different languages). 
These results are shown in Table~\ref{analysis-bt-st}. 

\begin{table}
\center
\begin{tabular}{|l|ccc|}
\hline
Model & Acc & $\Delta G$ & $\Delta R$ \\ \hline
Baseline & 75.5 & 0.4 & 18.8\\
ST$_{Gender}$ & 85.4 & \textbf{-4.4} & \textbf{-0.3}\\
BT$_{Gender}$ & \textbf{87.7} & -4.6 & 2.3\\
\hline
\end{tabular}
\caption{WinoMT performance for \textit{en-de} for the baseline and models using self-training data (ST$_{Gender}$) and back-translated data (BT$_{Gender}$).}
\label{analysis-bt-st}
\end{table}

The results highlight the flexibility of the gender-filtered data augmentation approach, as it can be applied to both source and target monolingual data. 
The results for BT are better overall than for self-training, whereas self-training is more convenient: 
We had to start with a much larger corpus in order to obtain the same amount of filtered data for BT as for self-training (90M vs.\ 26M sentences). 
In addition, since our focus was on translation from English into languages with grammatical gender, we were able to use the same filtered source monolingual data for all language pairs in the self-training approach. 
Finally, if using a fine-tuning paradigm (section~\ref{sec:abl-ft}), we do not even need to train a separate model for forward translation.

\subsection{Target Morphological Filtering}
\label{sec:filter-analysis}

\begin{table*}[ht]
\begin{tabular}{ll}
\multicolumn{2}{l}{\textbf{Incorrectly target-filtered sentences}} \\ \hline
fem & \textbf{She} had \textbf{her} \underline{share} of sorrows that money could not comfort. \\
    & Sie hatte ihren \underline{Anteil} an den Sorgen, die das Geld nicht tr{\"o}sten konnte. \\
\\
msc & \textbf{He} said: `I would give \textbf{him} a \underline{job} for life, but this is football. \\
    & Er sagte: ``Ich w{\"u}rde ihm {eine} {lebenslange} \underline{Arbeit} geben, aber das ist Fu{\ss}ball.\\
\end{tabular}
\caption{Example sentences incorrectly removed from the \textit{en-de} self-training corpus during target filtering (false negatives). The sentences are removed because there is a word in the target with the wrong grammatical gender (which is \underline{underlined} along with its aligned source word), even though in both cases this word is an inanimate noun. Note that the source sentences passed the source filtering step due to the gendered words in \textbf{bold}.}
\label{filter-example-falsenegative}
\end{table*}

This section analyses the quality of the target morphological filtering. In order to reduce error propagation from our self-training method, we automatically remove the forward translations that do not correctly reflect the gender of the source using target filtering. 
This is done using a morphological tagger and removing \textbf{all} sentences from the feminine-specific corpus that contain a grammatically masculine word (and similarly for the masculine corpus).\footnote{For languages with a neuter gender (DE, RU), we do not filter sentences based on the presence of a neuter gender word.} 

Note that this approach conflates grammatical gender and natural gender, which means that sentences with grammatical gender marked on unrelated nouns might be filtered unnecessarily. 
Table~\ref{filter-example-falsenegative} shows two such examples, where the feminine sentence is removed because the translation contains the masculine noun \textit{Anteil} (\textit{share}), and the masculine sentence is removed because of the feminine noun \textit{Arbeit} (\textit{job}). 
However, with this approach, sentences with \textbf{incorrectly} gendered translations are unlikely to be included in the final pseudo-parallel corpus. 
Indeed, as shown in Table~\ref{experiments-datasizes}, after target filtering we keep only 2-25\% of sentences that were present in the source-filtered data. 
We consider this to be an acceptable trade-off for the purposes of our work: we prefer to keep in high-confidence sentences at the cost of filtering valid sentences so as to minimize error propagation. 

\begin{table}
\center
\begin{tabular}{|l|c|c|c|c|}
\hline
Subset & TP & {TN} & {FP} & {FN}\\ \hline
feminine & 6\% & 8\% & 0\% & 86\% \\
masculine & 4\% & 4\% & 0\% & 92\% \\
\hline
\end{tabular}
\caption{Percent of true positives and negatives (\textbf{TP} and \textbf{TN}) as well as false positives and negatives (\textbf{FP} and \textbf{FN}) resulting from target morphological filtering on the \textit{en-de} pseudo-parallel data.}
\label{analysis-morph-filter}
\end{table}

We ran a small corpus analysis to estimate the trade-offs of our morphological filtering method. 
We selected a random 100-sentence sample of the forward-translated \textit{en-de} data and annotated each sentence for whether the gender was preserved in the translation.\footnote{The annotations were done by the authors of the paper, not by language experts.} 
We then compared this to the outcome of the filtering in order to estimate the rate of false positives and false negatives coming from this method. 
These results are shown in Table~\ref{analysis-morph-filter}. 

As desired, we do not see any false positives coming from filtering, meaning that errors in gendered translation are unlikely to be propagated due to the self-training procedure. 
On the other hand, this does come at a trade-off, as most of the sentences in the sample were valid but filtered unnecessarily. 
However, this analysis was done on the language pair with the highest baseline gender translation accuracy (\textit{en-de}), meaning that the vast majority of the translations correctly reflected the gender of the source. 
For our lower-resource  language pairs, we hypothesize this aggressive filtering will be even more beneficial than for \textit{en-de}. 


\section{Related Work}
\paragraph{Gender Translation Accuracy in MT:}

A large body of work has addressed bias in natural language processing (NLP) generally and MT specifically, surveyed in \citet{blodgett-etal-2020-language, ws-2019-gender, gebnlp-2020-gender, savoldi2021gender, sun-etal-2019-mitigating}, among others. In MT, several papers address the topic of gender in the context of ambiguous input or propose methods to control for gender or to augment data with gender information \citep{elaraby2018gender, moryossef-etal-2019-filling, DBLP:journals/corr/abs-1809-02208, saunders-etal-2020-neural, stafanovics2020mitigating, vanmassenhove-etal-2018-getting}. This prior work differs from ours in that we instead address the problem of gender accuracy for unambiguous inputs through gender balancing techniques. 

Work addressing the gender data imbalance issue in NLP~\cite{zhao-etal-2018-gender} is most related to this paper, as we take a similar approach inspired by data imbalance and propose self-training methods for augmenting gender-specific data. 
In machine translation, \citet{saunders-byrne-2020-reducing} showed that gender translation accuracy for unambiguous inputs can be improved through fine-tuning on small gender-balanced but counterfactual data. 
Specifically, their work proposed an approach that extracts a subset of source sentences containing gender-specific words (e.g.\ \textit{woman}, \textit{she}) and changes the gender of these words (e.g.\ \textit{man}, \textit{he}). The subsequent translations are used to create a dataset that is used to fine-tune the original model. Similar approaches are taken in \citep{costa-jussa-de-jorge-2020-fine,marcus:eit2021}. 
Unlike counterfactual data augmentation, our method does not alter the source data according to specific patterns. It instead uses naturally occurring data that we filter for gender phenomena. Additionally, we require only monolingual data which increases the flexibility and performance of our approach.

Another popular approach to reducing gender bias in NLP has been to use techniques such as hard and soft debiasing of embeddings~\cite{bolukbasi2016}. 
In NMT, \citet{escude-font-costa-jussa-2019-equalizing} used pre-trained debiased word embeddings instead of the word embeddings learned along with the NMT model and found that hard-debiased embeddings improved gender accuracy. 
Since this approach is orthogonal to our data augmentation approach, we experimented with using hard-debiased word embeddings alongside the augmented data.

\paragraph{Self-Training for MT:}
Monolingual data has been exploited via self-training methods to enhance statistical MT~\cite{DBLP:conf/iwslt/Schwenk08,DBLP:conf/iwslt/Ueffing06} as well as neural MT~\cite{wu-etal-2019-exploiting} by either forward translation of source monolingual data~\cite{imamura-sumita-2018-nict,zhang-zong-2016-exploiting} or back-translation of target monolingual data~\cite{sennrich-etal-2016-improving}. 
We opted for self-training through forward translation in this work, following prior work that showed that unfiltered translation can be effective in NMT for model compression~\citep{kim-rush-2016-sequence}, non-autoregressive translation~\citep{zhou2020understanding}, and domain adaptation~\citep{currey-etal-2020-distilling}. 
We also add filtering and balancing to mitigate error propagation. 


\section{Conclusion}
In this paper, we have focused on improving gender translation accuracy for unambiguously gendered inputs. 
We have proposed a gender-filtered self-training approach as a means of creating additional gender-specific training data. 
This is done by filtering source monolingual data by gender, translating the data with a baseline model, and running a second round of filtering on the target translations. 
Using this additional data, our models have achieved large gains in gender translation accuracy without damaging overall translation quality. 
Interestingly, adding feminine-specific data does almost as well as adding balanced feminine and masculine data, possibly due to gender imbalance in the original training data. 

Although this paper has shown strong gains for gender translation accuracy on feminine and masculine inputs, there is room for improvement. 
In the future, we would like to expand our work to other genders and other language pairs. 
This expansion will not be trivial: the self-training aspect of our approach assumes that the initial model is good enough at gender translation, which might not be the case for more under-represented genders or lower-resource language pairs. 
The use of automatic morphological analysis for target-side filtering additionally limits the applicability of this approach to other genders or language pairs. 
Thus, we would like to explore alternatives to self-training, such as synthetic data generation, as well as other approaches to filtering, such as using round-trip translation~\cite{moon-etal-2020-revisiting}.

\section{Broader Impact}
\label{sec:impact}

This paper has presented an approach for reducing the gap in accuracy between masculine-referring and feminine-referring inputs. 
This work addresses potential representational harms that can come from bias against the feminine gender. 
We use only gender-marked \textit{words}, with gender marked either lexically (English) or morphologically (German, French, Hebrew, Italian, and Russian), as the basis for our definitions of feminine and masculine inputs. 
Thus, we do not use human subjects, ascribe gender to any specific person, or use gender as a variable in our work. 

This work has shown improvements in gender translation accuracy for translation from English into several relatively diverse languages. 
In addition, improvements on translation accuracy for feminine inputs do not harm overall translation quality or gender translation accuracy for masculine inputs. 
Our approach can easily be generalized to other source languages with only lexical gender (e.g.\ Chinese) and to other target languages with grammatical gender (e.g.\ Hindi). 
While our technique does not completely close the gap between accuracy on masculine and feminine inputs, it does significantly improve over the baselines and as such it is a step in the right direction. 

Relying exclusively on the WinoMT benchmark may give practitioners and users false confidence about the level of gender bias in their machine translation systems. While the method proposed in this paper uses a generic monolingual corpus as the basis for our gender-specific data, our evaluation is limited to the available benchmarks: WinoMT and MuST-SHE. 
In order to mitigate the risk of overfitting to a specific benchmark, we have included human evaluations of accuracy and quality in addition to the standard automatic evaluations. 
However, given the availability of evaluation data for this task, we are not able to thoroughly test if the method proposed introduces other biases with respect to gender or other protected groups. For future work, we plan to expand on evaluation benchmarks and also use any additional benchmarks that may become available to the community.

This paper has only considered two genders (masculine and feminine). The proposed self-training approach relies on the baseline model being able to correctly translate the under-represented gender (in this case, feminine) for at least some inputs. This assumption likely does not hold for other under-represented genders, at least for the commonly used machine translation datasets. Additionally, the filtering step relies on a morphological analyzer to detect grammatical gender of the target words, which may not be straightforward for non-binary genders. Finally, although the WinoMT dataset we use for evaluation covers neutral gender, it does not cover non-binary gender, making this difficult to evaluate. In the future, we plan to expand our work towards covering other genders by creating additional evaluation benchmarks. 

\section*{Acknowledgments}
We would like to thank Margo Lynch, Tanya Badeka, Sony Trenous, and Felix Hieber for their help in evaluations. 
We would also like to thank the anonymous reviewers for their feedback. 

\bibliography{acl_latex}
\bibliographystyle{acl_natbib}

\clearpage
\newpage
\appendix
\section{Data, Preprocessing, and Hyperparameters}
\label{sec:app-params}
\paragraph{Parallel Data:} 
We train \textit{en-fr} on the WMT14 news task~\citep{bojar-etal-2014-findings}, \textit{en-it} on the IWSLT13 task~\citep{cettolo2013report}, \textit{en-ru} on WMT16~\citep{bojar-etal-2016-findings}\footnote{We only use Common Crawl, News Commentary v11 and Wiki Headlines corpora for training as we were not able to download the Yandex Corpus.}, \textit{en-he} on IWSLT14~\citep{cettolo2014report}, and \textit{en-de} on WMT18~\citep{bojar-etal-2018-findings}\footnote{Consistent with \citet{edunov-etal-2018-understanding}, we exclude the ParaCrawl corpus.}. 
For each language pair, we use the standard validation and test sets from the corresponding shared task. 

\paragraph{Monolingual Data:} 
For gender-filtered self-training, we use English News Crawl 2017 as the monolingual source data for all five language pairs. 
To balance the larger \textit{en-fr} parallel corpus, we also obtain feminine samples from English News Crawl 2015 and 2016 for that language pair. 

For target-side filtering, we use the spaCy morphological analyzer\footnote{\url{https://spacy.io/}} for FR and IT, pymorphy2~\citep{pymorphy} for RU, German-morph-dictionary based on DeMorphy~\citep{altinok2018demorphy} for DE and character-based rules following \citet{stanovsky-etal-2019-evaluating} for HE.

\paragraph{Preprocessing:} 
For all language pairs, we follow \citet{edunov-etal-2018-understanding} by removing sentences with more than 250 words or with a source/target length ratio higher than 1.5. 
We tokenize the data using the Moses tokenizer~\citep{koehn-etal-2007-moses}. 
We learn shared BPE vocabularies~\citep{sennrich-etal-2016-neural} with 32K types for DE and IT, and 40K types for FR. 
For RU and HE, we learn separate BPEs for source and target, source with 32K types for both and target with 2K types for HE and 32K types for RU. 

We use all the extracted feminine sentence pairs, and an equal number of masculine sentence pairs, during self-training for all languages except IT, where due to the small parallel data size we pick 30K random pairs. 
Similarly, due to the large size of the \textit{en-fr} parallel corpus, we up-sample the gender-specific pseudo-parallel data twenty times for that language pair. 

\paragraph{Training:}
We adopt training hyperparameters from \cite{edunov-etal-2018-understanding,ott-etal-2018-scaling}, and use the $\mathit{transformer\_wmt\_en\_de\_big}$ architecture with dropout rate \cite{srivastava2014dropout} of 0.3 for \textit{en-de/he/it/ru}, and dropout rate of 0.1 for \textit{en-fr}. 
We use the Adam optimizer \cite{kingma2014adam} with $\beta_1$=0.9, $\beta_2$=0.92 and $\epsilon$=1e-8, learning rate scheduler proposed by \citet{vaswani2017attention}, label smoothing ($\epsilon$=0.1) with uniform prior, and learning rate warm up for the first 4000 steps when training models. 
We use learning rate of 1e-3 for training \textit{en-de} models and for all other language pairs we use learning rate of 5e-4. 
Baseline \textit{en-de} and \textit{en-fr} models are trained for 30K and 180K\footnote{The number of updates are enough for all models to reach convergence in terms of validation perplexity.} synchronous updates respectively. 
During self-training, we increase the number of updates in proportion to the number of new samples added. 
For the other three language pairs, with relatively smaller training data sizes, we stop training when validation perplexity does not improve for 5 consecutive epochs. All models are trained on Nvidia V100 GPUs with 16-bit floating points precision, with parameter update frequency adjusted to simulate 64 GPUs training for \textit{en-de/fr} and 8 GPUs training for the other three language pairs. 
Final models are obtained through stochastic averaging of last 10 checkpoints.

\begin{table*}[ht]
\small
\center
\begin{tabular}{|l|ccc|ccc|ccc|ccc|ccc|}
\hline
&  \multicolumn{3}{|c|}{\textit{en-de}} & \multicolumn{3}{c|}{\textit{en-fr}} & \multicolumn{3}{c|}{\textit{en-he}} & \multicolumn{3}{c|}{\textit{en-it}}  & \multicolumn{3}{c|}{\textit{en-ru}} \\ 
Model & Fem & Msc & $\Delta S$ & Fem & Msc & $\Delta S$ & Fem & Msc & $\Delta S$ & Fem & Msc & $\Delta S$ & Fem & Msc & $\Delta S$\\     \hline
Baseline & 
    78.0 & 78.4 & \textbf{4.1} & 70.9 & 70.7 & 16.9 & 41.0 & 54.8 & 27.0 & 23.3 & 54.8 & 13.9 & 19.5 & 52.2 & 1.5\\
ST$_{Rand}$ & 
    82.5 & 80.9 & 4.5 & 68.9 & 69.5 & 15.2 & 42.0 & 54.5 & 26.2 & 21.2 & 55.9 & \textbf{11.0} & 19.2 & 51.4 & \textbf{-1.2}\\ 
\hline
ST$_{Gender}$ & 
    \textbf{90.8} & \textbf{86.4} & 4.5 & \textbf{76.7} & \textbf{75.4} & \textbf{9.3} & \textbf{42.1} & \textbf{55.9} & \textbf{23.6} & \textbf{45.8} & \textbf{63.8} & 12.1 & \textbf{26.4} & \textbf{56.7} & \textbf{1.2}\\
\hline
\end{tabular}
\caption{Additional WinoMT metrics not shown in Table~\ref{result-winomt-main} for the baseline, ST$_{Rand}$ baseline, and our ST$_{Gender}$ model. We show $F_1$ score on feminine inputs (\textbf{Fem}) and masculine inputs (\textbf{Msc}) as well as $\Delta S$ score.}
\label{result-winomt-more}
\end{table*}

\section{Full WinoMT Results}
Table~\ref{result-winomt-more} shows additional metrics on the WinoMT test set that were not shown in Section~\ref{sec:results}. 
Specifically, we show the $F_1$ scores on masculine and feminine inputs, as well as $\Delta S$. 
We examine the gender-specific $F_1$ scores to ensure that gains from our proposed ST$_{Gender}$ model do not harm any specific gender, and indeed we see that our ST$_{Gender}$ model achieves higher $F_1$ than both baselines for all language pairs and both genders studied. 
Our models do not specifically address stereotypicalness, and $\Delta S$ scores of our models are comparable to those of the baselines, indicating that our models do not exacerbate stereotype-related bias issues.

\section{Results on Generic Test Sets for Single-Gender Models}
\label{sec:generic-test}

In this section, we show BLEU scores on the generic test sets for the single-gender models in section~\ref{sec:abl}. 
Table~\ref{result-bleu-single} shows that the single-gender (feminine-only or masculine-only) data augmentation performs similarly to the baseline and to the model augmented with feminine and masculine data. 

\begin{table}
\small
\center
\begin{tabular}{|l|ccccc|}
\hline 
Model & \textit{en-de}  & \textit{en-fr} & \textit{en-he} &  \textit{en-it} & \textit{en-ru}\\     \hline
Baseline & 41.7 & \textbf{40.5} & 23.4 & 34.5 & 25.7\\
ST$_{Gender}$ & \textbf{41.8} & 40.2 & 23.8 & \textbf{34.6} & \textbf{26.6} \\
\hline
ST$_{Fem}$  & 41.7 & 40.3 & \textbf{24.0} & 34.3 & 25.8 \\ 
ST$_{Msc}$ & 41.7 & 40.4 & 23.2 & 34.3 & 26.5 \\ \hline
\end{tabular}
\caption{BLEU scores on the baseline model and on self-trained models. ST$_{Fem}$ uses only the feminine-specific data for augmentation, while ST$_{Msc}$ uses only the masculine-specific data.}
\label{result-bleu-single}
\end{table}

\end{document}